%% file: main.tex
\documentclass[conference]{IEEEtran}
\IEEEoverridecommandlockouts
\usepackage{lipsum}
\usepackage{graphicx}
\usepackage{xcolor}
\usepackage{datetime}
\usepackage{subcaption}
\usepackage{cite}
\usepackage{amsmath,amssymb,amsfonts}
\usepackage{algorithmic}
\usepackage{graphicx}
\usepackage{textcomp}
\usepackage{xcolor}
\usepackage{hyperref}
\def\BibTeX{{\rm B\kern-.05em{\sc i\kern-.025em b}\kern-.08em
    T\kern-.1667em\lower.7ex\hbox{E}\kern-.125emX}}
\begin{document}

	\title{Autonomous Multirotor Landing on Landing Pads and Lava Flows}

	\author{
		\IEEEauthorblockN{Joshua Springer}
		\IEEEauthorblockA{\textit{Department of Computer Science (Reykjavik University)} \\
		Supervised by: Marcel Kyas\\
		Reykjavik, Iceland \\
		joshua19@ru.is}
	}

	\maketitle

	\begin{abstract}
		\input{sections/abstract.tex}
	\end{abstract}

	\section{Introduction: Why Autonomous Landing?}
	\input{sections/introduction.tex}

	\section{Progress So Far}
	\subsection{Fiducial System Tests}
	\input{sections/fiducial_system_tests.tex}

	\subsection{Real World Tests}
	\input{sections/real_world_tests.tex}

	\subsection{Drone Platforms}
	\input{sections/drone_platforms.tex}

	\section{The Plan Moving Forward}

	\subsection{RAVEN and Holuhraun Lava Flow Landings}
	\input{sections/raven_holuhraun}

	\subsection{Onboard Depth Image Processing}
	\input{sections/onboard_depth_image_processing.tex}

	\subsection{Testing in Simulation}
	\input{sections/simulation.tex}

	\subsection{Local Analog Tests}
	\input{sections/local_analog_tests.tex}

	\subsection{Unified Sensor Set: Landing Pads + Lava Flows}
	\input{sections/unification.tex}

	\bibliography{main}
	\bibliographystyle{plain}


\end{document}

%% file: sections/abstract.tex
Landing is a challenging part of autonomous drone flight and a great research opportunity.
This PhD proposes to improve on fiducial autonomous landing algorithms by making them more flexible.
Further, it leverages its location, Iceland,
to develop a method for landing on lava flows
in cooperation with analog Mars exploration missions taking place in Iceland now
-- and potentially for future Mars landings.

%% file: sections/introduction.tex
Landing is a critical part of multirotor drone flight that has not yet been fully automated.
While some solutions exist, they are typically blind to the environment around the landing site,
or they are too computationally expensive to run onboard the drone and therefore require external
infrastructure.
The most widespread method of autonomous drone landing is for the drone to navigate to a specific
waypoint and descend vertically until it reaches the ground, not dynamically considering obstacles
or terrain hazards.
Some drones augment this with simple obstacle detection using, for example, ultrasonic sensors.
Importantly, GPS alone is not sufficiently accurate in many conditions
(e.g.~bad weather, urban canyons, even certain areas of the world),
with positional errors sometimes reaching multiple meters~\cite{gps_error_estimation}.
Real Time Kinematic (RTK) base stations can reduce the GPS error to the centimeter level,
but must be stationary near the drone's operational site,
requiring pre-mission setup, power, and limiting the range of the drone.

Another paradigm is to use a priori knowledge of known landing sites.
Some landing sites actively guide the drone to a landing using infrared emitters~\cite{precision_land_website},
or by recognizing the drone with a camera and relaying a command to it --
again requiring extra power and external infrastructure.
Fiducial markers are a popular way of passively marking landing pads for computationally cheap
visual pose estimation with a monocular RGB camera~\cite{fiducial_landing_two_fixed_cameras_apriltag,fiducial_landing_downward_facing_90_deg_gimbaled_camera,fiducial_landing_many_markers_voting_fixed_camera,fiducial_landing_ship_6dof_single_fixed_downfront_camera_apriltag,fiducial_vessel_landing_ar_tag_two_fixed_cameras}.
This often requires no extra sensors on the drone
because monocular RGB cameras are a very common peripheral drone sensor.
Many methods of fiducial autonomous landing use cameras that are either rigidly mounted to the drone
and pointed vertically downward or at some offset angle,
or mounted on a gimbal for stabilization, but pointing directly down.
These simplify the coordinate system transforms for the pose estimation,
because a weakness of fiducial markers is that their orientations are subject to ambiguity.

A third paradigm is to analyze the terrain beneath the drone in order to determine whether it is
safe enough to land, without having seen it before attempting the landing.
These methods often use LIDAR sensors for high-resolution terrain scans~\cite{conv_3d_lidar_landing,rooftop_landing},
and then process the data either on sophisticated hardware onboard the drone,
or offload the data to ground stations for processing (incurring data compression and transfer overhead).
In the past, the hardware available to process the data often required too much power
for small-scale drones, or was too slow to run in real time.
With the advent of embedded GPUs and TPUs, however, we have a chance to embed such processing into
many drones.

%% file: sections/fiducial_system_tests.tex
\label{section:fiducial_system_tests}
Starting at the master thesis level~\cite{joshua_master_thesis},
the author carried out autonomous landings in Gazebo Simulator
on the default Iris quadcopter running ArduPilot.
Using a hybrid arrangement of one WhyCon marker and one April Tag marker, the drone could search its
environment for a landing pad and track it with its gimbal-mounted camera,
estimate its relative position, and carry out autonomous landings.
For this purpose, relative positions are referred to in the East, North, Up (ENU) coordinate frame,
where ``East'' corresponds to the drone's right, ``North'' to the front, and ``Up'' to the space above the drone.
While the fiducial-based pose estimation worked most of the time,
the drone would occasionally perceive the landing pad to be in the wrong place,
e.g.~if the landing pad were at $\mathrm{ENU}=(1,2,3)$, then the drone could mistake it for
$(-1, 2, 3)$ or $(1, -2, 3)$.
This is because of the aforementioned orientation ambiguity that is inherent to fiducial markers.
The first part of the PhD was to study this phenomenon in several fiducial systems in order to determine
which has the least prevalence of orientation ambiguity,
and to test if those can run in real time on a Raspberry Pi 4 to be embedded on a drone.
The author has tested 2 unmodified systems: April Tag 48h12 and WhyCode,
and created/tested 3 variants of those systems.
The first variant, ``WhyCode Ellipse,'' shows significantly reduced orientation ambiguity as compared
to the original WhyCode system, while maintaining the high detection rate (~25 Hz).
The second variant ``WhyCode Multi'' regresses a plane to multiple presumed-coplanar WhyCode markers,
as their position estimates have low error, and takes the orientation of the plane to be the orientation
of all of the markers.
This system introduces complications that decrease the detection rate
and provide no decrease in orientation ambiguity.
The third variant, ``April Tag 24h10'' was designed to increase the detection rate of April Tag 48h12 on
a Raspberry Pi 3 B+, which was approximately 2 Hz in our experiments,
as a result of a large hash table that April Tag loads into memory.
Although it does provide a speed boost on the Raspberry Pi 3, the Raspberry Pi 4 has more computational
resources and therefore can handle both April Tag 24h10 and 48h12.
April Tag 24h10 also has high orientation ambiguity.
Detailed results have been submitted to IRC~\cite{fiducial_paper}.

\begin{figure}[]
    \centering
    \begin{subfigure}[b]{0.45\linewidth}
        \includegraphics[width=\textwidth]{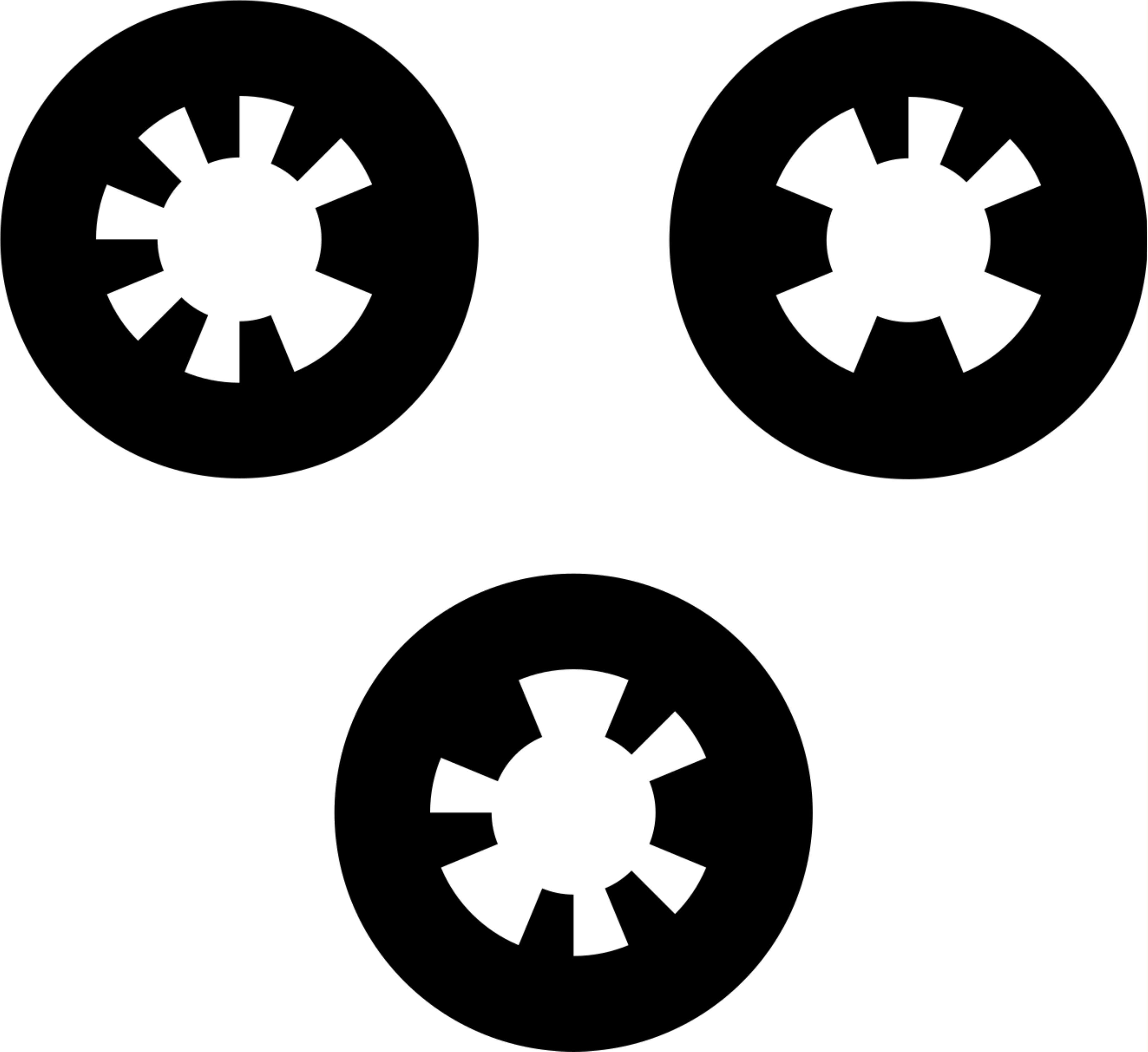}
        \caption{WhyCode ``Bundle''}
        \label{figure:whycode_bundle}
    \end{subfigure}
    \begin{subfigure}[b]{0.45\linewidth}
        \includegraphics[width=\textwidth]{./images/tagCustom24h10_00002_00001_00000.pdf}
        \caption{April Tag 24h10}
        \label{figure:apriltag24h10}
    \end{subfigure}

    \begin{subfigure}[b]{0.45\linewidth}
        \includegraphics[width=\textwidth]{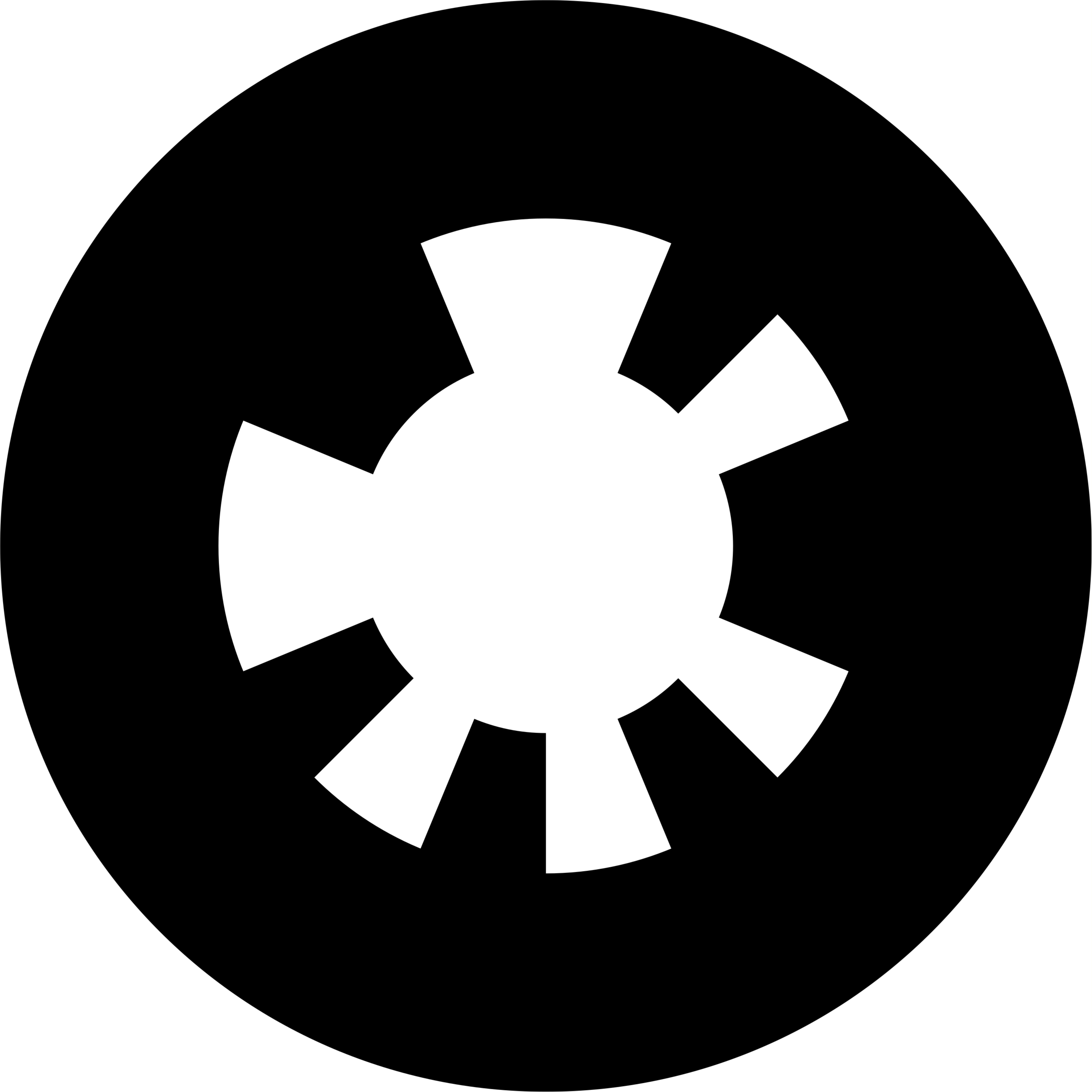}
        \caption{WhyCode}
        \label{figure:whycode_single}
    \end{subfigure}
    \begin{subfigure}[b]{0.45\linewidth}
        \includegraphics[width=\textwidth]{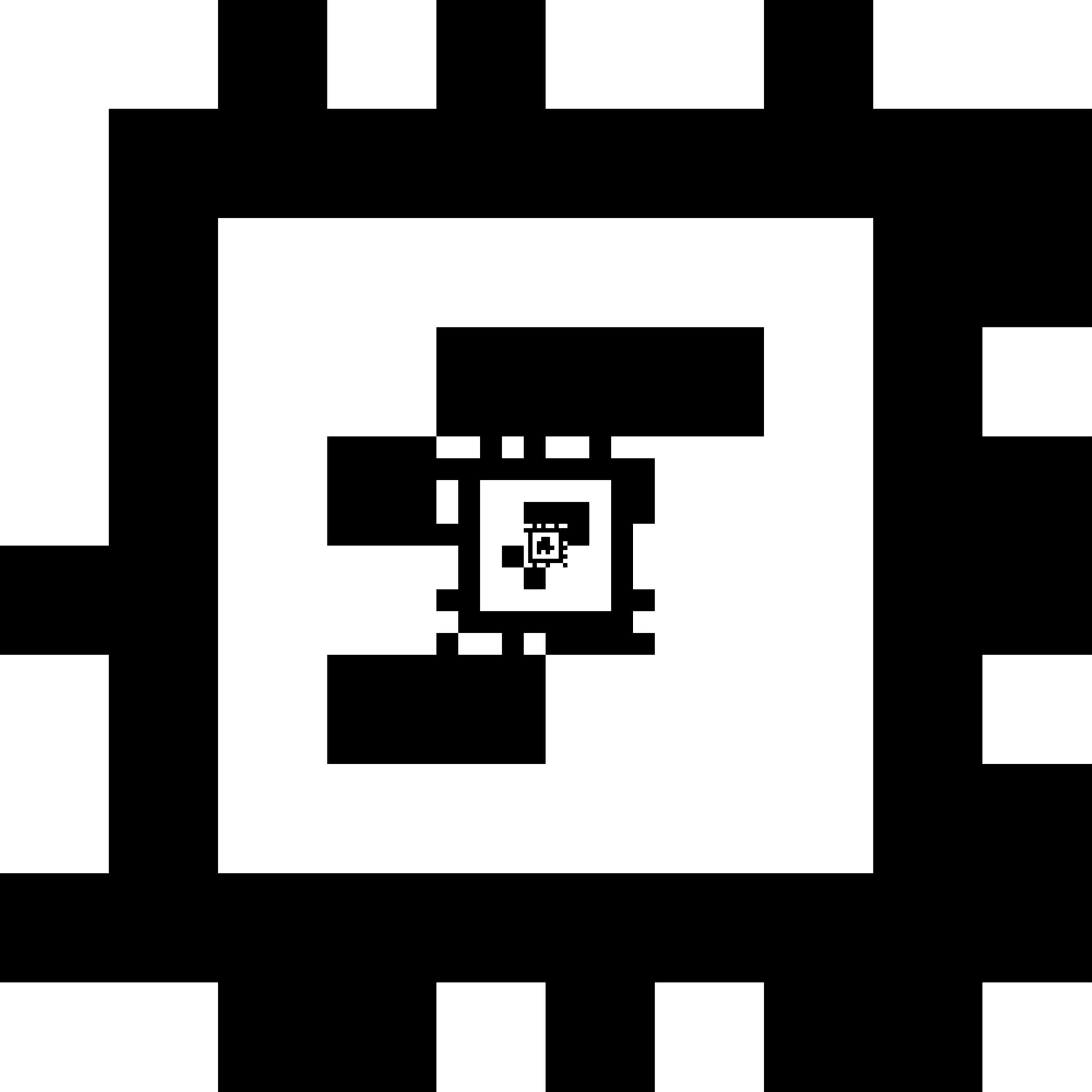}
        \caption{April Tag 48h12}
        \label{figure:apriltag48h12}
    \end{subfigure}
	\caption{The 4 landing pads tested with 5 fiducial systems. (2 systems use the WhyCode marker.)}
    \label{figure:marker_setup}
\end{figure}

%% file: sections/real_world_tests.tex
\begin{figure}
	\centering
	\includegraphics[width=\linewidth]{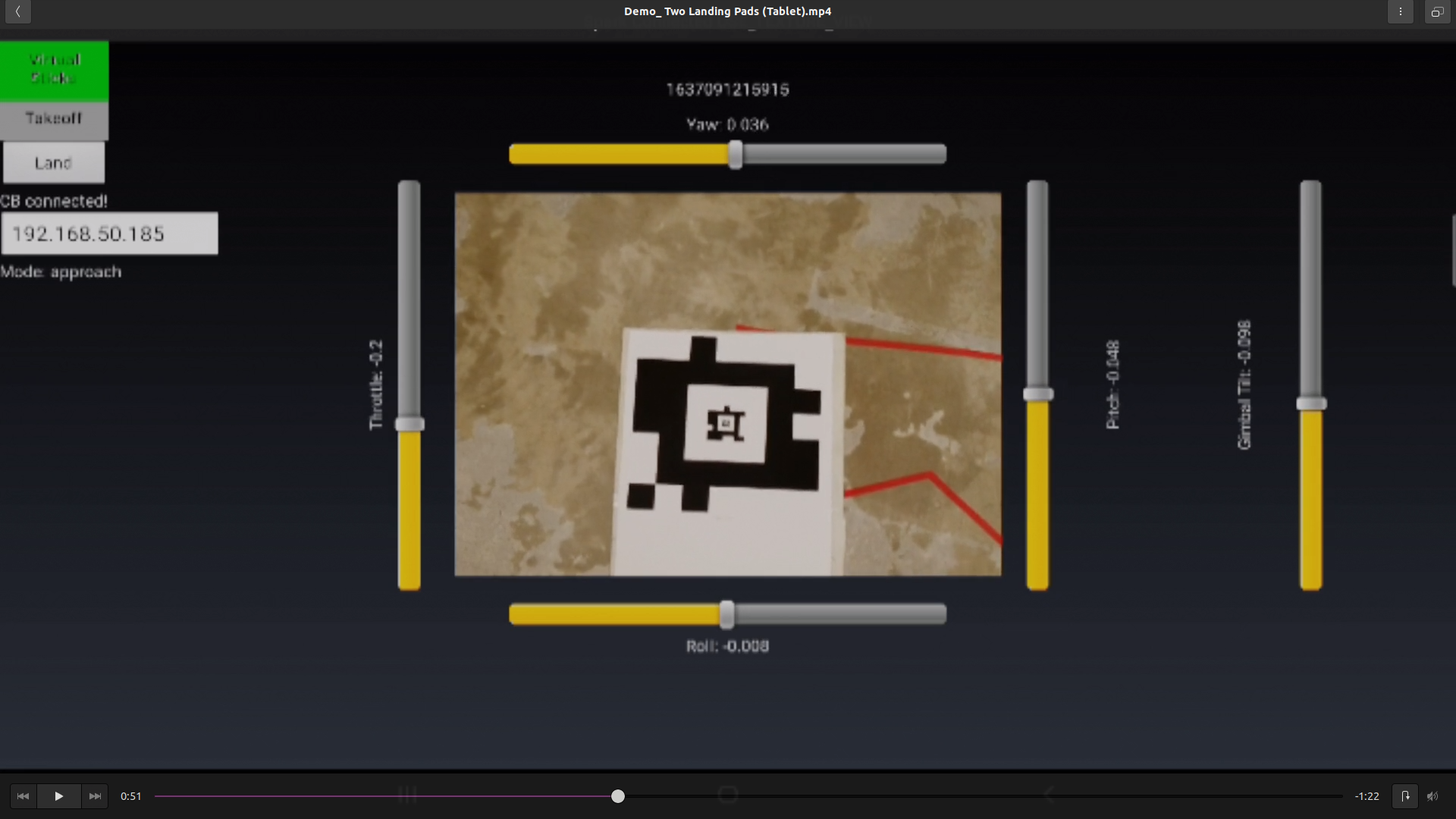}
	\caption{The app for the DJI Spark landing tests. Demo video: \url{https://vimeo.com/664863992}}
	\label{figure:spark_screenshot}
\end{figure}

\label{section:real_world_tests}
The next step was to test the fiducial systems again
in real world autonomous landing tests with a DJI spark.
While the Spark provides a small, sophisticated platform that allows for indoor testing during the
harsh Icelandic winter, it does not have onboard computational hardware that is available to the user.
However, it is possible to offload the video feed to a Raspberry Pi for processing,
via the controller and then an app on a tablet (see Figure~\ref{figure:spark_screenshot}) --
a workaround that causes some delay in video transmission but allowed the tests to move forward.
The Raspberry Pi generates VirtualStick commands that act as velocity setpoints in order to 
direct the drone during approach and landing.
4 of 5 fiducial systems produced successful landings on a $1\mathrm{m}^2$ landing pad,
ultimately showing that the orientation ambiguity is not a prohibitive obstacle,
provided that the drone has a safety-oriented control policy with slow, conservative movements.
The drone reliably locates the landing pads by simply spinning in place and tilting its gimbal-mounted
camera up and down, and then approaches the landing pad, descends, touches down, and disables the motors
without human intervention.
Detailed results have been submitted to IRC~\cite{fiducial_landing_paper}.

%% file: sections/drone_platforms.tex
Throughout the PhD, the author has built and flown several drones -- notably 2 Tarot 680 hexacopters with
sophisticated companion boards for onboard processing beyond fiducial systems.
One has an embedded TPU (Google Coral) and the other a GPU (NVIDIA Jetson Nano).
The flight system is a Pixhawk Cube Orange running PX4~\cite{px4}, with a Here 3 GPS that provides adequate
positioning accuracy for autonomous flight in Iceland (where GDOP tends to be high),
and an Ark Flow (optical flow for x/y velocity + LIDAR rangefinder for altitude and z velocity)
sensor for eventual GPS-denied autonomous flight.
The main peripheral sensor is an Intel RealSense D455 stereo depth camera that can provide high quality
terrain representations at 15 FPS to the companion boards.
We also have a heavy-lift hexacopter for carrying larger instruments, and have successfully deployed
it over the volcano in Fagradalsfjall in 2021 to collect aerial IR video with a FLIR A65
(featured on BBC Click: \url{https://www.bbc.co.uk/programmes/p09r8nzv}).
These have been tested successfully in autonomous surveys with real time, first-person video
from the depth camera.
This drone hardware will facilitate the anticipated experiments moving forward.

%% file: sections/raven_holuhraun.tex
RAVEN (Rover-Aerial Vehicle Exploration Network)
is a NASA-supported, Planetary Science and Technology from Analog Research (PSTAR)
project working to test collaborative drone/rover teams for Mars exploration~\cite{raven_website}.
They use Iceland -- specifically the Holuhraun lava flow (from 2014),
near the volcano Askja -- as a test bed for their analog research.
One goal of the project is to land an autonomous drone on a solidified lava flow,
to take geological samples such as cores, hyperspectral measurements, laser spectrography, and more.
Although the drone operations team can select landing sites that are likely to be viable
using images of Mars,
the drone itself must be able to guarantee that it has reached a safe location with a finer resolution
than is available from such images.
This landing scenario presents a very interesting challenge: the drone must be able to identify
safe landing sites that it has never seen, in a very rough and unforgiving environment --
and it must do this \textit{very} quickly and with very little power.
The maximum flight time for drones on Mars is currently about 6 minutes
because of limitations on power and size.
Further, the processing itself (and the power supplied to the sensors)
must be lightweight to be able to run both in real time and with no
support external to the drone.
This provides a collaborative path forward for the PhD,
and the author has gotten insight into the problem
by participating in RAVEN's field work during July and August of 2022.

%% file: sections/onboard_depth_image_processing.tex
\begin{figure}[]
    \centering
    \begin{subfigure}[b]{0.49\linewidth}
        \includegraphics[width=\textwidth]{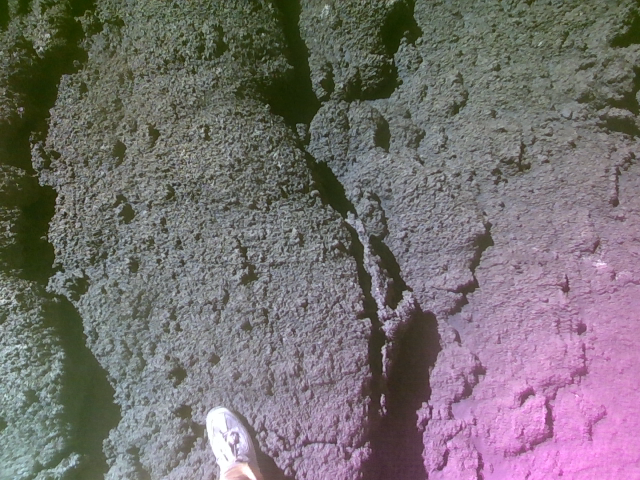}
	    \caption{Viable landing site (RGB).}
        \label{figure:viable_rgb}
    \end{subfigure}
    \begin{subfigure}[b]{0.49\linewidth}
	\includegraphics[width=\textwidth]{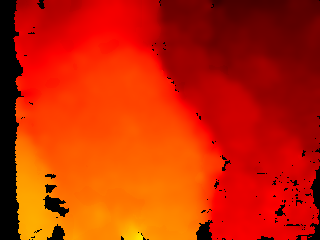}
	    \caption{Viable landing site (depth).}
        \label{figure:viable_depth}
    \end{subfigure}
    \begin{subfigure}[b]{0.49\linewidth}
        \includegraphics[width=\textwidth]{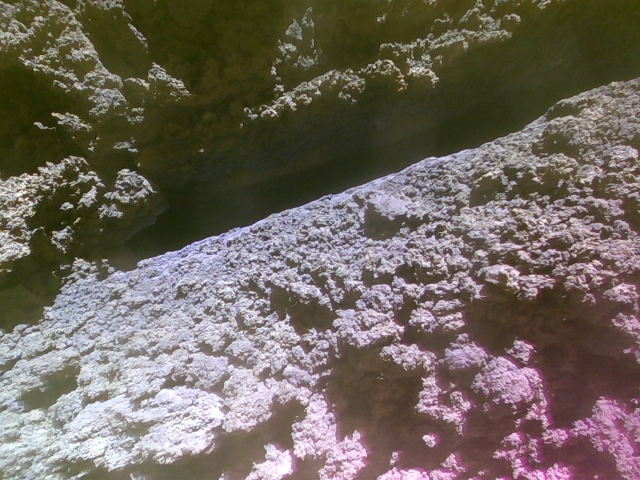}
	    \caption{Lava flow crack (RGB).}
        \label{figure:lava_flow_crack_rgb}
    \end{subfigure}
    \begin{subfigure}[b]{0.49\linewidth}
	\includegraphics[width=\textwidth]{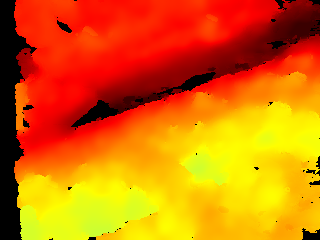}
	    \caption{Lava flow crack (depth).}
        \label{figure:lava_flow_crack_depth}
    \end{subfigure}
    \begin{subfigure}[b]{0.49\linewidth}
        \includegraphics[width=\textwidth]{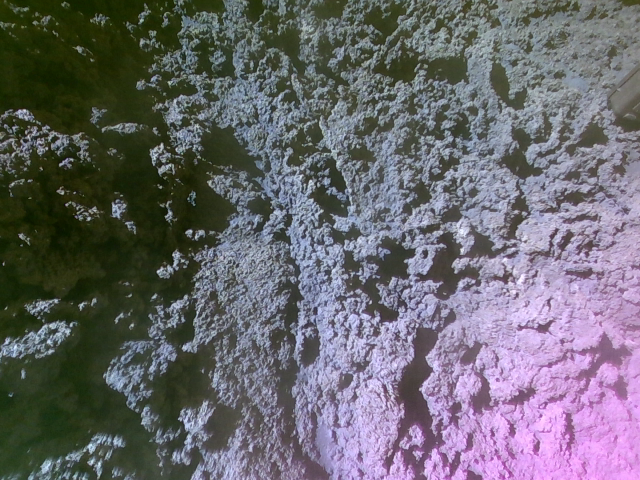}
	    \caption{Lava rough surface (RGB).}
        \label{figure:lava_flow_rough_surface_rgb}
    \end{subfigure}
    \begin{subfigure}[b]{0.49\linewidth}
	\includegraphics[width=\textwidth]{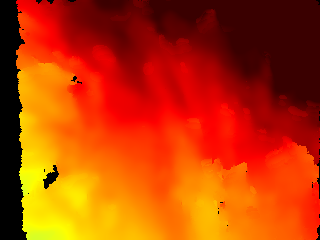}
	    \caption{Lava rough surface (depth).}
        \label{figure:lava_flow_rough_surface_depth}
    \end{subfigure}
    \begin{subfigure}[b]{0.49\linewidth}
        \includegraphics[width=\textwidth]{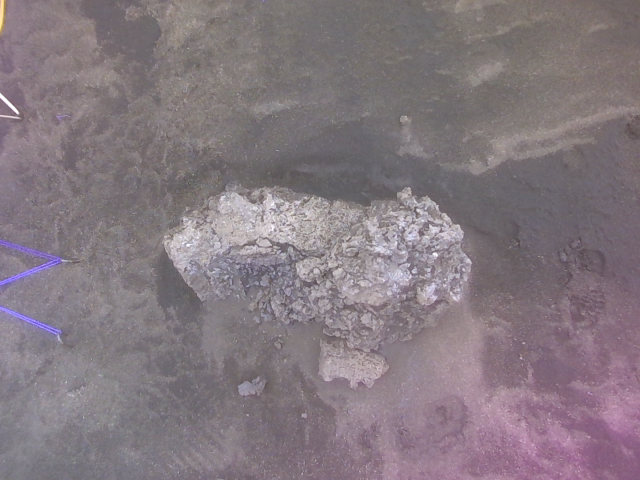}
	    \caption{Boulder obstacle (RGB).}
        \label{figure:boulder_obstacle_rgb}
    \end{subfigure}
    \begin{subfigure}[b]{0.49\linewidth}
	\includegraphics[width=\textwidth]{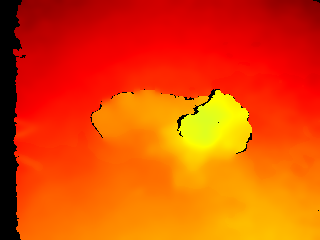}
	    \caption{Boulder obstacle (depth).}
        \label{figure:boulder_obstacle_depth}
    \end{subfigure}
    \caption{Various RGB and depth images from the Holuhraun lava flow,
	showing the surfaces and textures that the drone must land on or avoid.}
	\label{figure:rgb_and_depth_images}
\end{figure}

During the 2022 fieldwork, the author collected some terrain data via depth images
with a RealSense D455 depth camera --
both with it suspended from the heavy lift drone, and handheld while
walking through the lava flow.
The camera uses stereo images to generate a depth image of the scene in front of it,
where each pixel corresponds to a distance from the camera.
It also has an inertial measurement unit (IMU) that outputs its orientation with respect to gravity,
such that the depth images can be oriented parallel to the ground
(so that the orientation of the camera does not have to be fixed).
Figure~\ref{figure:rgb_and_depth_images} shows its performance
in the future test environment (darker colors represent longer distances).

The drone will need to locate sufficiently large, smooth, level areas in lava flows.
The most likely such candidates are on lava rise plateaus in pāhoehoe-like lava flows,
which tend to have unbroken and level surfaces,
such as the viable landing sites in Figures~\ref{figure:viable_rgb} and~\ref{figure:viable_depth}
~\cite{lava_texture_types}.
However, the drone will still need to detect and avoid uneven surfaces and cracks
as in Figures~\ref{figure:lava_flow_crack_rgb} and~\ref{figure:lava_flow_crack_depth}.
Similarly, it must be able to detect prohibitively rough surfaces that are not ideal for landing,
such as in
Figures~\ref{figure:lava_flow_rough_surface_rgb} and~\ref{figure:lava_flow_rough_surface_depth},
and salient obstacles such as the boulder in Figures~\ref{figure:boulder_obstacle_rgb}
and~\ref{figure:boulder_obstacle_depth}.
The simple absence of obstacles should not constitute a safe landing site in the absence of a
positive detection of a viable area.
This matters in cases of undefined (black) regions of the depth images,
where the disparity is too high for a depth estimate, such as in the lava crack,
or around the boulder.
One challenge is to fill in such holes that originate not from obstacles but rather from noise,
which could be solved by developing a probabilistic map of the area using many depth images,
with the assumption that noisy undefined regions will not persist over time.
The system will dilate obstacles and rough areas to give some margin of safety during the landing,
then select landing sites based on the centroid of the closest, sufficiently large, safe region.

After selecting a landing site, the drone could approach and land using the method
of velocity targets in Section~\ref{section:real_world_tests}, with the position of the selected
landing site as input instead of the position of a fiducial marker.
However, since the area will be more dangerous than a flat landing pad,
the drone should descend more quickly in the late stages of landing,
when the ground effect makes it harder to maintain a horizontal position.

Further interesting solutions could include more sophisticated techniques.
For example, the author would also like to explore the idea of locating safe regions specifically
for the shape of the landing gear, such that the drone could potentially straddle objects or cracks.
Deep learning techniques could play a role in combining RGB and depth images,
for example through semantic segmentation of the RGB images with a U-net~\cite{terrain_segmentation_roughness},
or through some unsupervised method that would not require a labeled data set.
Our onboard hardware -- the Google Coral TPU and Jetson Nano GPU --
are optimized for such edge computing and could therefore execute such techniques efficiently.
The main challenge in such a scenario would
be in curating the large amount of required data.

Successful methods will be tested for power consumption and runtime processing rate similarly
to the fiducial system tests from Section~\ref{section:fiducial_system_tests}.
Adequately fast methods will be deployed in autonomous 
landing missions in Holuhraun in upcoming fieldwork sessions.

%% file: sections/simulation.tex
Simulators provide an ideal environment for initial testing of autonomous drone control algorithms
by reducing logistical considerations,
which is critical for conducting research during the Icelandic winter.
AirSim~\cite{airsim} is a high-fidelity simulator into which one can import digital terrain models,
such as those taken by the RAVEN team at Holuhraun.
It also provides a drone model, integration with PX4 firmware, and configurable depth cameras
to help develop drone control methods.
AirSim has also been used to generate synthetic data for training deep learning networks,
and its ability to generate segmentation masks for RGB images can be particularly useful
for creating large labeled data sets.
If these synthetic data sets come from digital terrain models of the areas near the target landing site,
it may be possible to generate sufficiently realistic synthetic data sets to allow a
deep learning network
to semantically segment the terrain, and use the depth and RGB images together.
This is the only feasible way to generate a data set for this task in the timeline of the PhD,
as hand labeling such a data set would be prohibitively time-expensive.

%% file: sections/local_analog_tests.tex
\begin{figure}
	\centering
	\includegraphics[width=0.9\linewidth]{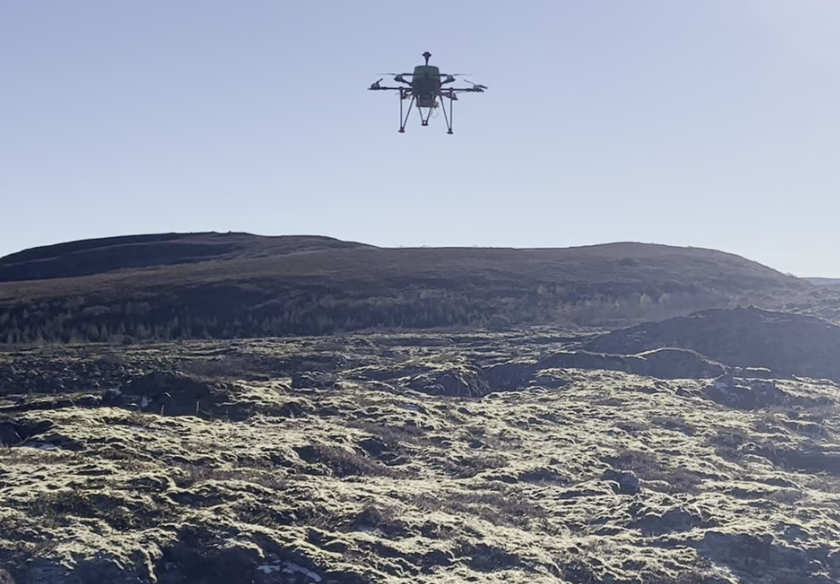}
	\caption{Collecting data in the Stóra-Bollahraun lava flow.}
	\label{figure:bollahraun}
\end{figure}

Iceland has many lava fields,
some of which are close to Reykjavik Univeristy
and therefore suitable for regular analog missions.
One such lava flow is Stóra-Bollahraun (see Figure~\ref{figure:bollahraun}),
which, although it has different textures from Holuhraun,
serves as a point for real world data collection
GPS-denied flight tests,
and landing tests
before the massive undertaking of further fieldwork in the highlands.
Here, the author has collected more data similar to that in Figure~\ref{figure:rgb_and_depth_images},
in order to a larger test set to process during the winter.

%% file: sections/unification.tex
The final proof of concept will ideally include a GPS-denied, autonomous mission
similar to the Spark's mission from Section~\ref{section:real_world_tests}.
The drone will take off from a marked landing pad
head towards a known lava flow to locate a landing site and conduct a landing,
take off again,
and then fly back to land at its original landing pad.
The D455 will provide the necessary sensor data to run both the fiducial landing algorithms
and the lava flow landing algorithms.
The fiducial algorithm will not even need to consider the unreliable orientation
of the markers as before, but can rather eliminate them by using the camera's reliable orientation
estimate.